\documentclass[runningheads]{llncs}

\usepackage{eccv}

\usepackage{eccvabbrv}

\usepackage{graphicx}
\usepackage{booktabs}

\usepackage[accsupp]{axessibility}  

\usepackage[pagebackref,breaklinks,colorlinks,citecolor=eccvblue]{hyperref}

\usepackage{orcidlink}
\usepackage{url}
\usepackage{multirow}
\usepackage{graphicx} 
\usepackage{booktabs} 
\newcommand{\tabincell}[2]{\begin{tabular}{@{}#1@{}}#2\end{tabular}}

\begin{document}

\title{From CLIP to DINO: Visual Encoders Shout in Multi-modal Large Language Models} 

\titlerunning{From CLIP to DINO: Visual Encoders Shout in MLLMs}

\vspace{-0.3cm}
\author{Dongsheng Jiang$^{1}$\thanks{Equal contribution. This work was done when Yuchen Liu and Jin Li worked as interns at Huawei Cloud.}, Yuchen Liu$^{2*}$, Songlin Liu$^{1*}$, Jin'e Zhao$^{3}$, Hao Zhang$^{3}$, Zhen Gao$^{3}$, Xiaopeng Zhang$^{1}$, Jin Li$^{2}$, Hongkai Xiong$^{2}$.}

\authorrunning{D. Jiang et al.}

\institute{$^1$Huawei Cloud \quad
$^2$Shanghai Jiao Tong University \quad
$^3$Yunding Technology \\ {\tt\small dongsheng\_jiang@outlook.com, wspolsl@gmail.com, zxphistory@gmail.com}\\
{\tt\small\{liuyuchen6666, deserve\_lj, xionghongkai\}@sjtu.edu.cn}}

\maketitle

\vspace{-0.55cm}
\begin{abstract}
  Multi-modal Large Language Models (MLLMs) have made significant strides in expanding the capabilities of Large Language Models (LLMs) through the incorporation of visual perception interfaces. Despite the emergence of exciting applications and the availability of diverse instruction tuning data, existing approaches often rely on CLIP or its variants as the visual branch, and merely extract features from the deep layers. However, these methods lack a comprehensive analysis of the visual encoders in MLLMs. In this paper, we conduct an extensive investigation into the effectiveness of different vision encoders within MLLMs. Our findings reveal that the shallow layer features of CLIP offer particular advantages for fine-grained tasks such as grounding and region understanding. Surprisingly, the vision-only model DINO, which is not pretrained with text-image alignment, demonstrates promising performance as a visual branch within MLLMs. By simply equipping it with an MLP layer for alignment, DINO surpasses CLIP in fine-grained related perception tasks. Building upon these observations, we propose a simple yet effective feature merging strategy, named \textbf{COMM}, that integrates \textbf{C}LIP and DIN\textbf{O} with \textbf{M}ulti-level features \textbf{M}erging, to enhance the visual capabilities of MLLMs. We evaluate \textbf{COMM} through comprehensive experiments on a wide range of benchmarks, including image captioning, visual question answering, visual grounding, and object hallucination. Experimental results demonstrate the superior performance of \textbf{COMM} compared to existing methods, showcasing its enhanced visual capabilities within MLLMs.
\end{abstract}

\vspace{-0.89cm}
\section{Introduction}
\vspace{-0.25cm}
\label{sec:intro}
Large Language Models (LLMs)~\cite{chatgpt, openai2023gpt4, touvron2023llama, llama2, taori2023stanford, chiangvicuna} have made significant strides in the domains of language understanding and generation, achieving remarkable progress recently. Through instruction tuning~\cite{wei2021finetuned,wang2022self}, existing LLMs demonstrate their versatility as general-purpose models capable of handling a wide range of tasks. This capability unlocks their potential zero-shot learning ability, enabling seamless task switching guided by instructions. Building upon the promising performance of LLMs, researchers are now motivated to enhance their capabilities by incorporating visual signals as inputs. This extension allows the generation of textual outputs that are closely related to visual content, opening up exciting possibilities in the realm of vision-language understanding.

To this end, Flamingo~\cite{alayrac2022flamingo} and BLIP2~\cite{li2023blip2} align the powerful LLMs with a frozen visual encoder to understand visual inputs and perform various vision-language tasks. A series of following works, LLaVA~\cite{liu2023llava}, InstructBLIP~\cite{dai2023instructblip}, MiniGPT-4~\cite{zhu2023minigpt} and mPLUG-OWL~\cite{ye2023mplug} further improve the ability to follow human instructions by constructing multi-modal instruction-following datasets for training. However, these methods are built on image-level alignments, which suffer from the limited fine-grained understanding (such as region description~\cite{liu2017referring} and reasoning~\cite{zellers2019recognition}) and severe object hallucination problem~\cite{li2023obj_Hallucination}. To this end, GPT4ROI~\cite{zhang2023gpt4roi} proposes instruction tuning on region-of-interest and unlocks the region-level multimodal capacities. Kosmos-2~\cite{peng2023kosmos} and Shikra~\cite{chen2023shikra} further integrate the grounding abilities into LLMs and unlock the referential ability in dialogue, \emph{i.e.}, enable the user to point to the object or region as input and the model responds with spatial coordinates of bounding boxes. Such grounding capacity can fulfill numerous vision-language tasks, which is a great progress in MLLMs.  

Despite a wide variety of exciting methods and applications, most of existing multi-modal LLMs employ CLIP~\cite{radford2021learning} or its variants ~\cite{sun2023eva} as the visual branch, where the features output from the deep layers (\emph{e.g.}, the penultimate layer) are usually employed as inputs to the language decoders. However, it still lacks analysis that: \emph{Whether using the Vanilla CLIP features as visual encoder is the best way for MLLMs?} Though the visual encoder of CLIP is apparently well aligned with the word embedding space by image-text contrastive learning, it fails to learn more detailed pixel-level information such as color and positioning due to the global supervision of image captions, which might hinder the fine-grained perception ability in MLLMs. Besides, existing MLLMs have quite unbalanced visual and language encoders (\emph{e.g.}, ViT-Large-300M vs. Vicuna-7B/13B). Since the language models have succeeded in scaling up the model size with progressively powerful language abilities, the short plate of the Buckets Effect for MLLMs lies in the visual models, which fails to demonstrate emerging capabilities, and suffer from domain gap and limited zero-shot ability. Consequently, it is critical to enhance the visual capabilities for boosting MLLMs. 

This paper presents an extensive investigation into different visual encoders for MLLMs. Four typical visual foundation models are considered, \emph{i.e.}, image-text contrastive learning CLIP, image-only contrastive learning DINOv2~\cite{oquab2023dinov2}, masked image modeling MAE~\cite{he2022masked} and supervised learning DeiT~\cite{touvron2021training}. We evaluate the performance on commonly-used vision-language tasks including visual grounding, object hallucination, visual question answering, image captioning and MME benchmark. 
Our analysis reveals that different layers of features exhibit varying biases towards local and global patterns. Shallow layer features containing low-level detailed information prove beneficial for fine-grained perception tasks such as grounding and positioning ability, while deep layer features are superior at global understanding. To enhance representation, we propose a multi-level feature merging strategy that incorporates both low-level and high-level features. Surprisingly, when equipped with an MLP layer for alignment, the vision-only model DINOv2 shows promise as a visual branch for MLLMs. We attribute this to the fine-grained localization information captured by DINOv2. Conversely, MAE and DeiT perform inferiorly as visual branches for MLLMs. MAE learns limited semantic information, while DeiT's strong supervised training makes the alignment with the textual space challenging. Based on the above observations, we propose a fusion strategy that integrates \textbf{C}LIP and DIN\textbf{O} with \textbf{M}ulti-level features \textbf{M}erging), dubbed as \textbf{COMM}, for boosting the visual branches of MLLMs.
Experimental results demonstrate clear advantages of the proposed model over existing approaches and highlight the enhanced visual capabilities brought by \textbf{COMM}. In a nutshell, the contributions of this paper are summarized as follows:
\begin{itemize}
    \item We are the first to extensively investigate the effectiveness of different visual encoders for MLLMs. Based on the analysis that shallow layer features contain low-level detailed information which is helpful for fine-grained tasks, we propose a multi-level feature fusion strategy to incorporate low-level and high-level features for improving representation.
    \item Our analysis indicates that vision-only DINOv2 achieves promising results in MLLMs with only an MLP layer for alignment. Considering fine-grained pixel information in DINOv2 and global semantic information in CLIP, we propose \textbf{COMM} to fuse the visual embeddings of these two models to enhance visual capabilities for boosting MLLMs.
    \item Extensive experiments on a wide range of tasks including visual grounding, referring expression generation, object hallucination, visual question answering and image captioning demonstrate the superiority of \textbf{COMM} over existing works.
\end{itemize}
\vspace{-0.5cm}
\section{Related Work}
\vspace{-0.2cm}
\textbf{Multi-modal Large Language Model.} LLMs~\cite{dai2019transformer,brown2020language} have garnered significant attention in both academia and industry due to their remarkable understanding and generative abilities. The success of LLMs has motivated researchers to explore the integration of vision into these models, leading to the development of powerful multi-modal LLMs (MLLMs). Flamingo~\cite{alayrac2022flamingo} employs a cross-attention module to extract visual contexts, which are concatenated with text token as input for LLMs. LLaVA~\cite{liu2023visual} and FROMAGe~\cite{koh2023grounding} leverage the vision encoder of CLIP to extract visual features, which is aligned to text features using a single linear layer and then input to LLMs. Models such as BLIP-2~\cite{li2023blip}, mPLUG-OWL~\cite{ye2023mplug}, MiniGPT-4~\cite{zhu2023minigpt} and InstructBLIP~\cite{dai2023instructblip} employ Q-former to extract text-aligned visual features for LLMs. Recently, some interesting works extend LLMs to image retrieval~\cite{koh2023grounding}, video understanding~\cite{zhang2023video}, audio~\cite{su2023pandagpt}, biomedical analysis~\cite{li2023llava}, control systems~\cite{driess2023palme}. 

In recent studies, there has been a growing interest in extending MLLMs to improve their fine-grained understanding abilities through region-level image-text alignment. Kosmos-2~\cite{peng2023kosmos} addresses this by constructing a large-scale dataset of grounded region-text pairs, enabling the integration of grounding abilities into LLMs. GPT4RoI~\cite{zhang2023gpt4roi} reformulates the bounding box as a spatial instruction format and extracts visual features based on region-of-interest, facilitating region-level multi-modal understanding. Shikra~\cite{chen2023shikra} proposes a unified model that handles spatial coordinates to possess referential abilities in dialogue contexts. Ferret~\cite{you2023ferret} and ViP-LLaVA~\cite{cai2023making} further extends with a broader range of free-form shapes for referring, including points, boxes, sketches and scribbles.
Additionally, Qwen~\cite{Qwen-VL} presents a set of MLLMs that demonstrate remarkable performance across various tasks. However, previous works have predominantly focused on extracting visual features solely from the last few layers of the CLIP model, resulting in an emphasis on global image properties. In this study, we draw attention to the fact that features extracted from shallower layers exhibit a stronger focus on localized properties, which we argue can be more potent in comprehending object locations and image details. Additionally, while CLIP primarily learns globally aligned features, advanced vision-alone models such as DINOv2 excel in capturing more fine-grained vision features. We posit that leveraging these fine-grained vision features can effectively enhance the capabilities of MLLMs, as demonstrated in our analysis. To further advance this line of research, we introduce a novel fusion module that expands and enhances the visual branches, thereby aiming to significantly improve the performance of MLLMs.

\textbf{Large Vision Foundation Model.} Recent progresses in training vision foundation models with large-scale image data focus on contrastive learning, masked image modeling and supervised training. For one thing, contrastive learning can be conducted in an image-only or image-text manner. DINOv2~\cite{oquab2023dinov2} pretrains the image encoder on large curated image data, which shows a superior understanding of object parts and scene geometry across image domains. Image-text contrastive learning as CLIP~\cite{radford2021learning} and EVA-CLIP~\cite{sun2023eva} employs the natural language as weak supervision to guide the learning of visual features. 
For another, BEiT~\cite{HangboBao2021BEiT} predicts discrete tokens based on a pre-trained image tokenizer while iBOT~\cite{zhou2021ibot} proposes an online image tokenizer. MAE~\cite{he2022masked} proposes a masked autoencoder for reconstructing image pixels. Besides, DeiT III~\cite{Touvron2022DeiTIR} proposes a training recipe to achieve promising performance.
Recent MLLMs employ the vision encoder of CLIP/EVA-CLIP without considering the properties of specific visual models. 
In this paper, we are the first to re-examine the effectiveness of existing visual models in MLLMs and propose a simple yet effective fusion strategy for boosting visual capabilities.
\vspace{-0.35cm}
\section{Analysis of the Visual Branch in MLLMs}
\begin{figure}[!t]
\renewcommand{\baselinestretch}{1.0}
\centering
\includegraphics[width=1\textwidth]{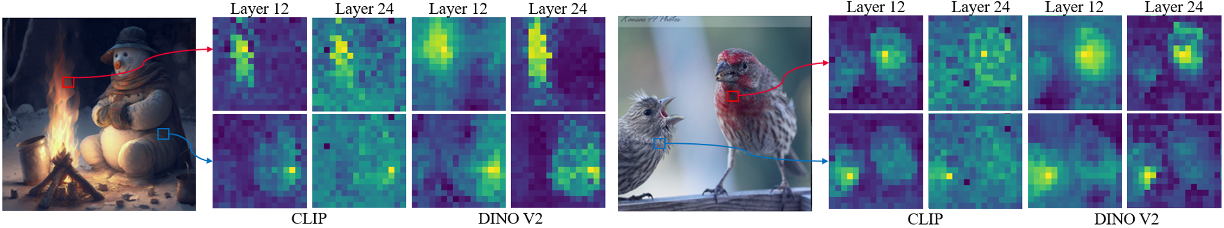}
\vspace{-16pt}
\caption{Feature correspondence visualization by computing the cosine similarity of different visual tokens extracted from the shallow and deep layers of CLIP and DINOv2. }\label{fig:feature}
\vspace{-10pt}
\end{figure}

\vspace{-0.2cm}
Previous MLLMs~\cite{liu2023llava,liu2023improvedllava,zhu2023minigpt,dai2023instructblip,ye2023mplug,peng2023kosmos,chen2023shikra,Qwen-VL,you2023ferret} usually utilize the vision encoder of CLIP as their visual branch. Typically, these models extract features from the last few layers, such as the penultimate layer, which are then fed into an alignment network. Subsequently, the aligned features are concatenated with text tokens to serve as input for the LLMs. While the image-text pretraining of CLIP aligns well with the language model, it primarily learns image-level features but overlooks the richer pixel-level features due to the constraint of limited fine-grained information in the guided captions. Moreover, the deep-layer features primarily focus on global image properties and inadequately explore the intricate details of local object parts. As depicted in Fig.~\ref{fig:feature}, the visual features extracted from the shallow layers of CLIP and the deep visual features obtained from the visual-only model DINOv2 contain more detailed information regarding local objects, such as shape or texture. Leveraging these detailed features may enhance the MLLMs' fine-grained perception abilities. 

\textbf{Evaluation Settings.} For further analysis, we conduct a series of quantitative experiments using different kinds of visual models, \emph{i.e.}, image-text contrastive learning CLIP, image-only contrastive learning DINOv2, masked image modeling MAE and supervised learning DeiT. In specific, the visual features extracted from different layers of visual models (based on ViT-Large) are aligned using a linear projection layer and then concatenated with text tokens as the input for LLMs (here we use Vicuna-7B~\cite{chiangvicuna}). The overall architecture and training process follow Shikra~\cite{chen2023shikra} but with fewer iterations (9400 iterations, batch size 16 on 4 A800) to save the computation cost. Then, we measure the capability of the trained MLLMs on referring expression comprehension (REC)~\cite{chen2023shikra}, referring expression generation (REG)~\cite{peng2023kosmos} and object hallucination benchmark (POPE)~\cite{li2023obj_Hallucination}. 
Detailed descriptions of these tasks can be referred to Sec.~\ref{sec:experiments}.

\begin{figure}[!t]
\renewcommand{\baselinestretch}{1.0}
\centering
 \subfloat[Average REC accuracy.]{
\includegraphics[width=0.33\columnwidth]{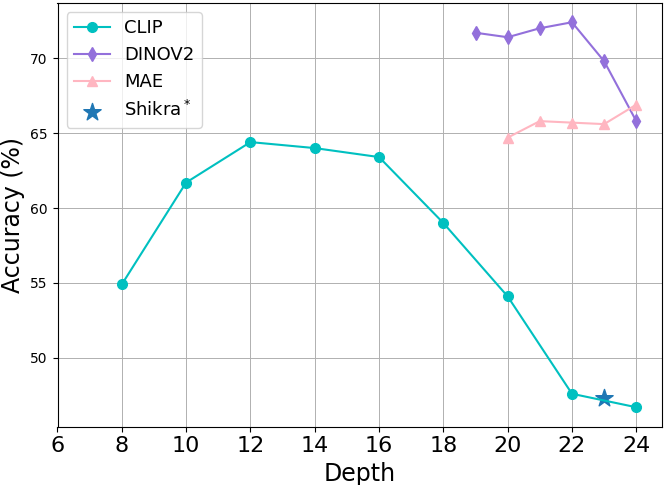}}
\subfloat[Average POPE accuracy.]{\includegraphics[width=0.33\columnwidth]{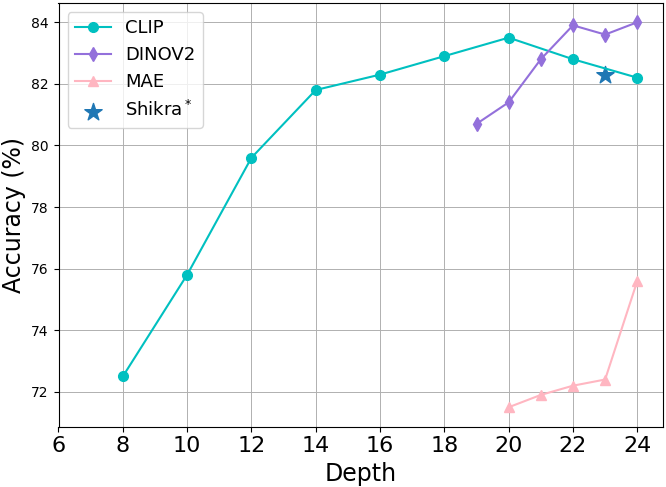}}
\subfloat[Average REG CIDEr.]{\includegraphics[width=0.33\columnwidth]{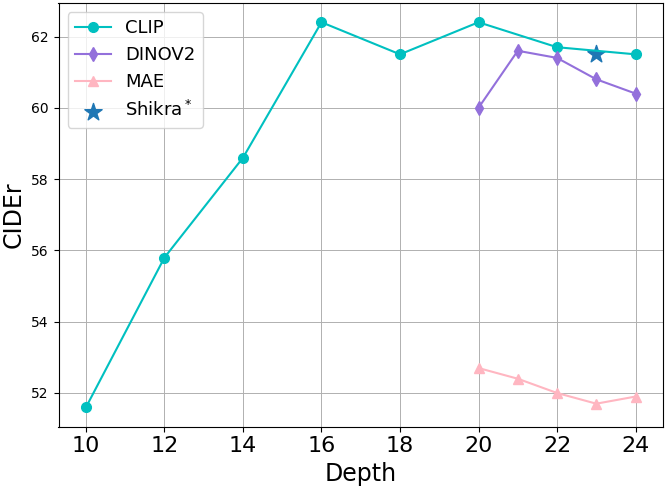}}
\vspace{-0.1cm}
\caption{Average REC, POPE accuracy and REG CIDEr for using different layers of features extracted from various vision models (CLIP, DINOv2 and MAE), as input to MLLMs. Shikra uses the 23rd layer features of CLIP and we reproduce its results with fewer iterations (denoted as Shikra$^*$).}
\label{fig:rec}
\vspace{-0.12cm}
\end{figure}

\textbf{CLIP as the Visual Branch of MLLMs.} As depicted in Fig.~\ref{fig:rec}, we observe that different layers of features exhibit varying biases towards grounding and understanding abilities. For instance, the shallow features demonstrate relatively higher accuracy in terms of REC and reach their optimal value at layer 12. Conversely, the deep features achieve higher accuracy in terms of POPE, indicating superior understanding ability. Notably, the relatively deep features (layer 16) display the best REG CIDEr score, showcasing promising region understanding capabilities. Consequently, instead of solely relying on deep features as done in previous works, we argue that integrating both shallow and deep features is crucial for MLLMs with improved overall performance.

We further explore various merging modes of low-level and high-level features. Denoting the output features from each transformer layer of ViT with a depth of $N$ as $\mathbf{z}=[ z_1,..,z_i,...,z_N ]$, we discuss several multi-level feature merging (MFM) strategies for combining shallow and deep features, namely: 

$\bullet$ \emph{Mean(half)}: averaging output patch token features in the second half of the backbone as $z=(z_{N/2}+\cdots+z_{N})/(N/2)$. 

$\bullet$ \emph{Mean(all)}: averaging features output by all layers as $z=(z_{1}+\cdots+z_{N})/N$. 

$\bullet$ \emph{Layerscale(all)}: learning a scale parameter as the weight to sum features output by all layers as $z=w_1z_{1}+\cdots+w_Nz_{N}$, where $w_i$ refers to the weight assigned to the $i$-th layer feature and all these weights are dynamically updated and summed up to 1. 

$\bullet$ \emph{LLN-Layerscale(all)}: using a linear-layernorm module to align the feature space between different layers’ features and then summed by \emph{Layerscale} as $z=w_1\mathrm{LLN}(z_{1})+\cdots+w_N\mathrm{LLN}(z_{N})$.

$\bullet$ \emph{Conv-Layerscale(all)}: using a convolution and bn module to align the feature space between different layers’ features and then summed by \emph{Layerscale} as $z=w_1\mathrm{Conv}(z_{1})+\cdots+w_N\mathrm{Conv}(z_{N})$. 

\begin{figure}[!t]
\renewcommand{\baselinestretch}{1.0}
\centering
 \subfloat[ REC acc for CLIP.]{
\includegraphics[width=0.26\columnwidth]{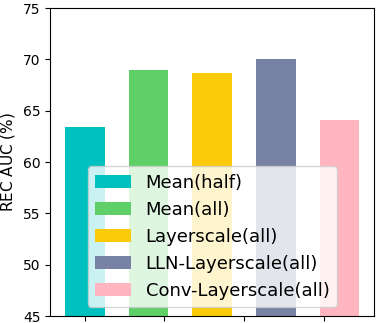}}
\subfloat[ POPE acc for CLIP.]{
\includegraphics[width=0.248\columnwidth]{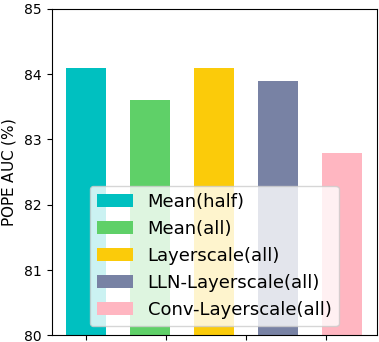}}
\subfloat[REC acc for DINO.]{\includegraphics[width=0.247\columnwidth,height=0.222\columnwidth]{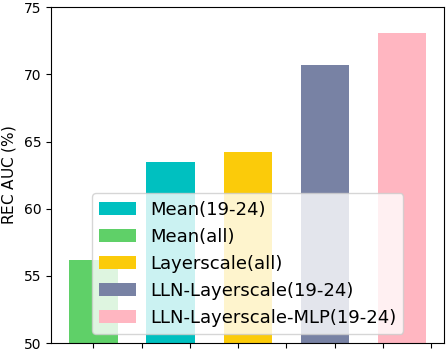}}
\subfloat[POPE acc for DINO.]{\includegraphics[width=0.248\columnwidth,height=0.222\columnwidth]{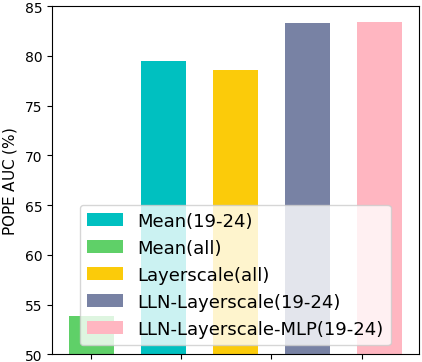}}
\vspace{-0.1cm}
\caption{Average REC and POPE accuracy for merging different layers of features with mutli-feature merging (MFM) strategies as input to MLLMs for visual backbones of CLIP and DINOv2. }
\vspace{-0.3cm}
\label{fig:clipanddino}
\end{figure}

Fig.~\ref{fig:clipanddino} (a) and (b) shows that simply averaging all shallow and deep features of CLIP can \emph{de facto} achieve a satisfactory accuracy and \emph{LLN-Layerscale} strategy further improves performance. With \emph{LLN-Layerscale} as MFM module, the performance of CLIP can be evidently improved on commonly-used vision-language tasks as shown in Table~\ref{tab:fusion}. 

\textbf{DINOv2 as the Visual Branch of MLLMs.} 
To leverage the rich fine-grained visual information present in DINOv2, but not inherently aligned with text, we employ a non-linear Multi-Layer Perceptron (MLP) module to align the image features with the word embedding space. Fig.~\ref{fig:rec}  demonstrates that the deep-layer features of DINOv2 exhibit superior grounding abilities, as evidenced by higher REC accuracy, and display satisfactory understanding abilities, as indicated by favorable POPE and REG results. Additionally, we explore the efficacy of multi-level feature merging to enhance performance. In contrast to CLIP, the merging of shallow features from DINOv2 leads to a significant performance degradation. Specifically, in Fig.~\ref{fig:clipanddino}(c) and (d), it is evident that \emph{Mean(all)} performs notably worse than \emph{Mean(19-24)} in terms of both REC and POPE accuracy, indicating that the shallow representations lack sufficient semantic information. Building upon the \emph{LLN-Layerscale} approach, the incorporation of the MLP module for a more potent connection between the visual and text spaces demonstrates a clear improvement in performance. Table~\ref{tab:fusion} showcases the substantial performance gains achieved by employing \emph{LLN-Layerscale-MLP} as Multi-Level Feature Merging (MFM) module across various vision language tasks. Further detailed ablation studies on the MLP module are in Section~\ref{sec:mae}. 
\begin{table}[!t]
\centering
\setlength{\tabcolsep}{2.8pt}
\caption{Comparison of the visual model using CLIP, DINOv2 with Multi-level Feature Merging (MFM) and \textbf{COMM} to incorporate visual embeddings of both models on VL tasks. CLIP baseline use the 23rd layer features, which follows Shikra but with fewer training iterations. DINOv2 baseline is w/o MLP module.
MME CS and PS indicate cognition and perception score, respectively.
}
\vspace{-0.1cm}
\label{tab:fusion}
\scalebox{0.82}{
\begin{tabular}{l|cccccccc}
\toprule
Visual Model  &Avg REC& Avg POPE &COCO & Flickr30k &MME CS&MME PS   & VQAv2&OK-VQA  \\
\cmidrule(lr){1-9}
CLIP&47.3&82.3&125.0&80.7&209.6&1107.8&68.8&44.2 \\
DINOv2&54.8&78.3&118.0&68.9&261.8&930.5&63.1&41.9\\
\cmidrule(lr){1-9}
CLIP w/ MFM&  70.0    & 83.4 & 125.8 & 81.0&296.6&1164.4 &69.5  & 44.7 \\
DINOv2 w/ MFM
& \textbf{72.8}&83.3&123.4&76.3 &  252.9 &1086.8&68.0 & 42.1 \\
 \cmidrule(lr){1-9}
\textbf{COMM}
& \textbf{72.8} & \textbf{83.6}&\textbf{127.3}&\textbf{81.9}  & \textbf{360.4}&  \textbf{1234.9}&  \textbf{70.1}& 	\textbf{45.0}\\
\bottomrule
\end{tabular}
}
\vspace{-0.26cm}
\end{table}

\textbf{MAE and DeiT as the Visual Branch of MLLMs.} Fig.~\ref{fig:rec} shows that MAE features achieve acceptable REC accuracy, but suffers large performance drop on POPE and REG evaluation. This is because MAE features lack sufficient semantic information for global or regional understanding. Therefore, MAE is not suitable as the visual branch for MLLMs. DeiT performs even worse than MAE (details in Section~\ref{sec:mae}). We speculate that this is because supervised training is too strong, which learns a specialized visual space that is difficult to align with the word embedding space.
\begin{figure}[!t]
\renewcommand{\baselinestretch}{1.0}
\centering
\includegraphics[width=1\columnwidth]{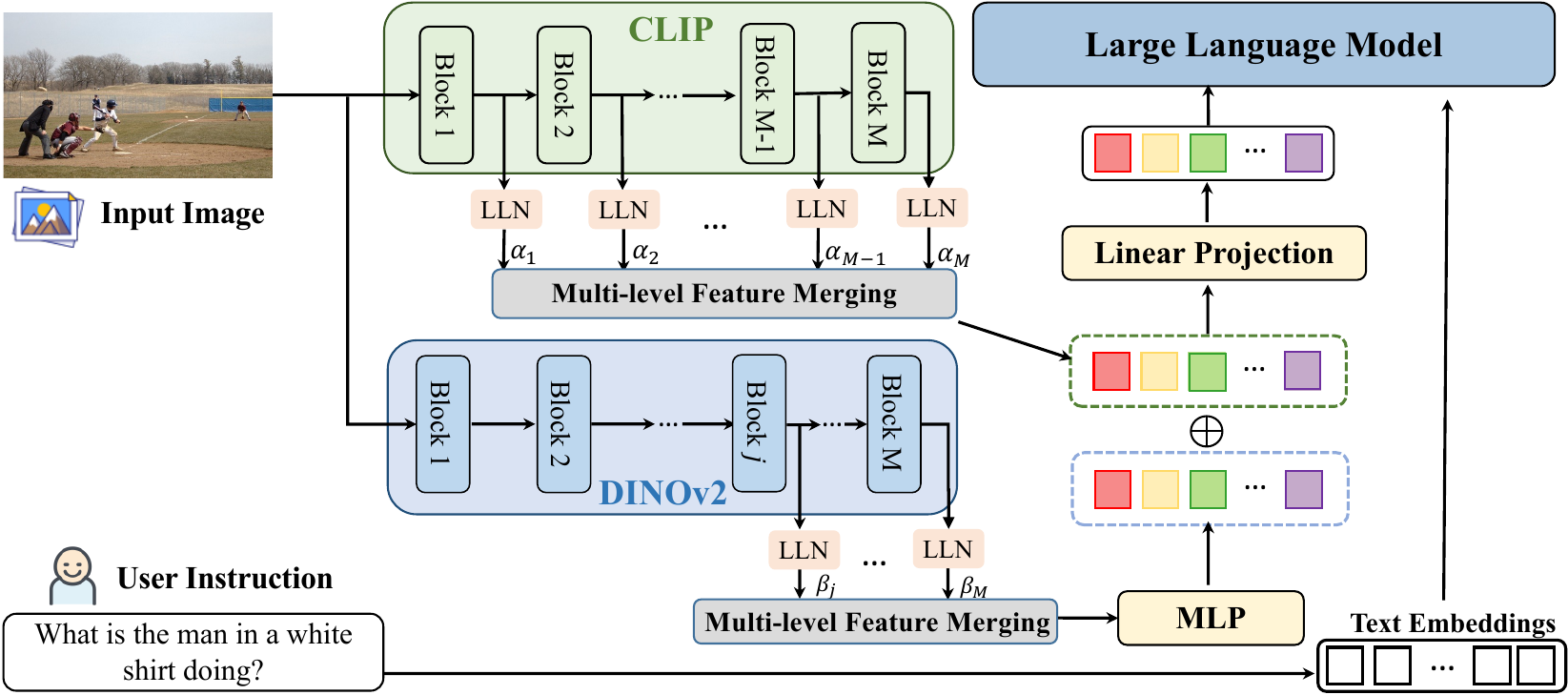}
\vspace{-0.29cm}
\caption{Overview of our \textbf{COMM}. The image is input to the vision encoder of CLIP and DINOv2, and the features from shallow and deep layers are incorporated by multi-level feature merging. The features of DINOv2 are aligned with an MLP and concatenated with features of CLIP, which are input to a linear layer. Then the fused features are concatenated with text tokens as input to LLMs.}
\label{fig:overall}
\vspace{-0.35cm}
\end{figure}

\section{\textbf{COMM}: Combining CLIP and DINO With Multi-level Feature Merging}
\textbf{Architecture Overview.} In this section, we introduce the proposed \textbf{COMM}, that integrates CLIP and DINO with Multi-level features Merging to enhance the visual capabilities of MLLMs. The overall framework is illustrated in Fig.~\ref{fig:overall}, \textbf{COMM} is incorporated into a vision-language instruction following model built upon the
recent advanced language and vision-language foundation models. Following the input instructions, our model takes vision and language as inputs to generate text responses following the input instructions.
Specifically, we adopt the visual encoder of CLIP and DINOv2 (based on ViT-Large) with our proposed fusion strategy as the visual branch, and Vicuna~\cite{chiangvicuna} (7B/13B) as language decoder. The visual encoder is downsampled with rate 14, meaning that an image with resolution $H\times W$ will be represented by a sequence of $\frac{H}{14}\times \frac{W}{14}$ tokens.
The fused token features are projected using a linear layer and then concatenated with the instruction tokens as inputs to the language decoder, which is a generic interface to unify various vision-language tasks as text generation task.

Specifically, denote the visual encoder of CLIP and DINOv2 (ViT Large used) as $f_1$ and $f_2$, respectively. Given an input image $x$, we extract the patch token features output by all layers of CLIP as $f_1(x)=[v_1^{1},...,v_1^{i},...,v_1^{24}]$, where $v_1^i \in \mathbf{R}^{N\times D}$, $N$ is the number of patch tokens and $D$ is the embedding dimension. The features output by the deep layers of DINOv2 are $f_2(x)=[v_2^{19},...,v_2^{i},...,v_2^{24}]$. Then we concatenate the features output by these two models as $\mathbf{v}=[v_1^{1},...,v_1^{24},v_2^{19},...,v_2^{24}]$. A linear-layernorm module is employed to align the feature space between different layers’ features and layerscale is used to merge multiple layer features as
\begin{equation}
    \overline{v}_1 = \sum_{i=1}^{24} \alpha_i \cdot \mathrm{Linear}(\mathrm{LN}(v_1^{i})), \qquad \overline{v}_2 = \sum_{j=19}^{24} \beta_j \cdot \mathrm{Linear}(\mathrm{LN}(v_2^{j}))
\end{equation}
where $\alpha$ and $\beta$ are the learnable scaling parameter. Then, we employ an MLP layer to project the features of DINOv2 and concatenate the output features with that of CLIP as $\mathbf{\overline{v}}=[\overline{v}_1,\mathrm{MLP}(\overline{v}_2)]$. Then, a linear layer is employed to match the dimension of visual features to that of text features as $\mathbf{\hat{v}}=\mathrm{Linear}(\mathbf{\overline{v}})$. Finally, fused visual features $\mathbf{\hat{v}}$ are concatenated with text tokens as input to LLMs.
\vspace{-0.1cm}
\section{Experiments}
\vspace{-0.1cm}
\label{sec:experiments}
In this section, we conduct extensive evaluation on four kinds of vision-language tasks to comprehensively evaluate the visual understanding ability of our model, namely, Referring Expression Comprehension, Referring Expression Generation, Object Hallucination Benchmark, and Visual Question Answering and Image Captioning.

\textbf{Training Details.} Similar to previous MLLM methods, \textbf{COMM} is trained in two stages. In the first pretraining stage, we train the model on the reorganized vision-language dataset as~\cite{chen2023shikra}, including public VQA, Image Captioning datset and several datasets containing positional annotation RefCOCO, visual gemone~\cite{krishna2017visualgenome} and Visual-7W~\cite{mani2020pointqa}. The first pretraining stage is conducted for 100K steps. In the second instruction tuning stage, we set the sampling ratio to 50\% on LLaVA-Instruct-150K~\cite{liu2023llava} and Shikra-RD~\cite{chen2023shikra}. Instead of 224 × 224 resolution currently used by existing MLLMs, we use 336 × 336 resolution to reduce the information loss caused by image down-sampling and promote the fine-grained perception ability. In both stages, we freeze the visual encoder and tune all parameters in LLMs, alignment layer and multi-level feature fusion module. We adopt AdamW~\cite{DBLP:conf/iclr/LoshchilovH19adamw} as the optimizer and cosine annealing
scheduler~\cite{DBLP:conf/iclr/LoshchilovH17cos} as learning rate scheduler with an initial learning rate of
2e-5 and global batch size of 64. All training runs on 8 NVIDIA A800 GPUs. It takes around 100h for stage one training and 20h for stage two.

\begin{table*}[!t]
\centering
\caption{Results on standard referring expression comprehension (REC) task. Generalist VL models can perform various vision-language tasks.
Specialist models are designed specifically for localization tasks or generalist pretraining models that undergone finetuning. The results of Shikra, Qwen, Ferret and Griffon are from their papers.
}
\vspace{-0.06cm}
\label{tab:rec}
\resizebox{\textwidth}{!}{%
\begin{tabular}{l|l|cccccccc}
\toprule
\multirow{2}{*}{Model type}
& \multirow{2}{*}{Model}  & \multicolumn{3}{c}{RefCOCO} & \multicolumn{3}{c}{RefCOCO+} & \multicolumn{2}{c}{RefCOCOg} \\
 &  & val & test-A & test-B & val & test-A & test-B & val-u & test-u \\
\cmidrule(lr){1-2}\cmidrule(lr){3-10}
\multirow{9}{*}{\tabincell{l}{Generalist VL SOTAs \\ (w/o finetuning)}}
& OFA-L*      & 79.96 & 83.67 & 76.39 & 68.29 & 76.00 & 61.75 & 67.57 & 67.58  \\
& VisionLLM-H   & - & 86.70     & - & - & - & - & - & - \\
& Shikra-7B      & 87.01 & 90.61 & 80.24 & 81.60 & 87.36 & 72.12 & 82.27 & 82.19 \\
& Shikra-13B     & 87.83 & 91.11 & 81.81 & 82.89 & 87.79 & 74.41 & 82.64 & 83.16  \\
&Ferret-7B&87.49&91.35&82.45&80.78&87.38&73.14&83.93&84.76\\
&Ferret-13B&89.48&92.41&84.36&82.81&88.14&75.17&85.83&86.34\\
&Griffon-13B&88.00&92.10&81.90&81.50&88.20&73.30&82.90&84.30\\
&Qwen-VL-7B&89.36& 92.26 &85.34 &83.12& 88.25& 77.21& 85.58& 85.48\\
&Qwen-VL-7B-Chat& 88.55& 92.27& 84.51 &82.82& 88.59 &76.79& 85.96 &86.32\\
& \textbf{COMM-7B (Ours)}     &\textbf{91.73}& \textbf{94.06}&\textbf{88.85}&\textbf{87.21}&\textbf{91.74}&\textbf{81.39}&\textbf{87.32}&\textbf{88.33}\\
\cmidrule(lr){1-2}\cmidrule(lr){3-10}
\multirow{3}{*}{\tabincell{l}{Specialist SOTAs \\ (Specialist/Finetuned)}}
& G-DINO-L     & 90.56 & 93.19 & 88.24 & 82.75 & 88.95 & 75.92 & 86.13 & 87.02 \\
& UNINEXT-H    & 92.64 & 94.33 & 91.46 & 85.24 & 89.63 & 79.79 & 88.73 & 89.37 \\
& ONE-PEACE    & 92.58 & 94.18 & 89.26 & 88.77 & 92.21 & 83.23 & 89.22 & 89.27 \\
\bottomrule
\end{tabular}%
}
\vspace{-0.2cm}
\end{table*}
\begin{table*}[!t]
\centering
\setlength{\tabcolsep}{4.5pt}
\caption{Results on standard referring expression generation (REG) task in CIDEr score. We reproduce the results of Shikra-7B using its officially released checkpoint. SLR is a finetuned listener-speaker model with an added reward-based module (SLR).
}
\vspace{-0.13cm}
\label{tab:reg}
\scalebox{0.95}{
\begin{tabular}{l|cccccccc}
\toprule
 \multirow{2}{*}{Model}  & \multicolumn{3}{c}{RefCOCO} & \multicolumn{3}{c}{RefCOCO+} & \multicolumn{2}{c}{RefCOCOg} \\
  & val & test-A & test-B & val & test-A & test-B & val-u & test-u \\
\cmidrule(lr){1-9}
SLR~\cite{yu2017joint}&-&69.7&132.3&-&49.4&70.9&59.2&-\\
SLR+Rerank~\cite{yu2017joint}&-&77.5&132.0&-&52.0&73.5&66.2&-\\
\cmidrule(lr){1-9}
Shikra     &75.61&44.26&104.83&56.42& 40.98&68.25&62.71&65.58\\
Kosmos-2    & - & - & - & - & - & - & 62.3 & - \\
\textbf{COMM} (Ours)   &\textbf{93.35}&\textbf{54.95}&\textbf{131.13}&\textbf{70.00}&\textbf{52.27}&\textbf{79.05}&\textbf{79.22}&\textbf{77.96}\\ 
\bottomrule
\end{tabular}}
\vspace{-0.05cm}
\end{table*}
\subsection{Referring Expression Comprehension}
To evaluate the fine-grained understanding and positioning capability of our model, we investigate the referring expression comprehension task on benchmarks as RefCOCO~\cite{kazemzadeh2014referitgame}, RefCOCO+~\cite{mao2016refcocog} and RefCOCOg~\cite{mao2016refcocog}, where models are asked to localize the object described with an expression. As shown in Table~\ref{tab:rec}, compared with generalist VL models and previous SOTA MLLMs, \textbf{COMM} achieves significant performance gain on all benchmarks, \emph{i.e.}, \textbf{COMM}-7B outperforms Shikra-13B and Qwen-VL-7B-Chat by 4.87\% and 3.10\% accuracy on average, respectively. With more powerful visual capabilities of our proposed fusion model, we can evidently surpass recent SOTA MLLMs in a more efficient way, \emph{e.g.}, using a smaller LLM than Shikra (7B vs. 13B) and less training data than Qwen (3.6M vs. 1.4B). Besides, our generalist model even achieves comparable results with specialist SOTA methods, showing the superior grounding ability of our MLLMs.

\subsection{Referring Expression Generation}
\vspace{-0.03cm}
Moreover, we evaluate the ability to understand image regions or objects referred via inputting bounding boxes. Instead of referring image regions or objects via detailed text descriptions, directly referring to image regions via its bounding boxes is more effective and can reduce the ambiguity.
The experiments are conducted on the referring expression generation task with RefCOCO, RefCOCO+ and RefCOCOg, aiming to generate text descriptions of specific regions in the bounding box. Table~\ref{tab:reg} shows that our model outperforms Shikra and Kosmos-2 by a considerable margin of 16.51 CIDEr and 16.92 CIDEr gain on RefCOCOg, demonstrating the effectiveness of our model for fine-grained understanding. Besides, \textbf{COMM} even outperforms finetuned SLR on RefCOCO+ and RefCOCOg.

\begin{table}[!t]
\centering
\setlength{\tabcolsep}{1.5pt}
\caption{Object hallucination benchmark using POPE evaluation pipeline \cite{li2023obj_Hallucination}. The results of Shikra-7B are taken from its paper. Except for Shikra-7B, the other results are obtained from \cite{li2023obj_Hallucination}.
}
\vspace{-0.06cm}
\label{tab:pope_results}
\scalebox{0.95}{
\begin{tabular}{l|ccccccc}
\toprule
Datasets  &\textbf{COMM}& Shikra & InstructBLIP  & MiniGPT-4 & LLaVA&MM-GPT & mPLUG-Owl \\
\cmidrule(lr){1-8}
Random
&   87.29   & 86.90 & \textbf{88.57} & 79.67 &50.37 &50.10 & 53.97 \\
Popular
&   \textbf{86.50}    & 83.97  & 82.77 &69.73  &49.87 &50.00 &50.90  \\
Adversarial
&   \textbf{84.50}    & 83.10  & 72.10  &65.17  &  49.70 &50.00& 50.67\\
\bottomrule
\end{tabular}%
}
\vspace{-0.3cm}
\end{table}
\begin{table*}[!t]
\centering
\renewcommand{\tabcolsep}{0.9mm}
\caption{Results on visual question answering (VQA) and image captioning.
For VQA, we evaluate SOTA generalist models and our \textbf{COMM} onVQAv2 and OK-VQA following the normalization rules. Shikra and LLaVA-1.5~\cite{liu2023improvedllava} is based on the 13B variant.
For image captioning, we evaluate them on COCO and Flickr30k in CIDEr score.
We call Flamingo as FM for short.
}
\vspace{-0.07cm}
\resizebox{\textwidth}{!}{%
\begin{tabular}{l|l|cccccccc}
\toprule
\multicolumn{2}{c|}{Datasets} &\textbf{COMM}&LLaVA-1.5&Qwen& Shikra  & FM-80B& BLIP-2 & Unified-IO  & VPGTrans\\
\cmidrule(lr){1-2}\cmidrule(lr){3-10}
\multirow{4}{*}{VQA}
& VQAv2$^\text{val}$     &\textbf{79.05}&-&-& 75.33  &   -   & 65.2 & -  & 65.2 \\
& VQAv2$^\text{dev}$     &\textbf{81.04}&80.0&79.5& 77.36  & 56.3 & 65.0 & 77.9  & - \\
& VQAv2$^\text{std}$     &\textbf{81.17}&-&-& 77.51  &  -  &  -  & -  & - \\
& OK-VQA                 &\textbf{59.18}&-&58.6& 47.16  & 50.6 & 45.9 & 54.0  & 45.0 \\
\cmidrule(lr){1-2}\cmidrule(lr){3-10}
\multirow{2}{*}{Caption}
& Flickr30k      &\textbf{88.2}&-&85.8 & 73.9 &  67.2   &   -  &   -   &   -   \\
& COCO           &\textbf{132.7}&-&- & 117.5 & 84.3  &   -  & 122.3 &   -  \\
\bottomrule
\end{tabular}%
}
\vspace{-0.5cm}
\label{tab:vl}
\end{table*}
\vspace{-0.1cm}
\subsection{Object Hallucination Benchmark}
We compare our model against the baseline models
on the hallucination evaluation dataset recently introduced by POPE~\cite{li2023obj_Hallucination}, which randomly selects 500 images from COCO~\cite{caesar2018coco}. Table~\ref{tab:pope_results} shows that \textbf{COMM} surpasses recent popular MLLMs with 1.44\% and 4.95\% higher accuracy on average than Shikra and InstrutBLIP, respectively. By enhancing the fine-grained visual capabilities, \textbf{COMM} can effectively alleviate the object hallucination problem.

\subsection{Visual Question Answering and Image Captioning}
We evaluate \textbf{COMM} on conventional VL tasks of VQA and Image Captioning. Specifically, image captioning requires the model to generate description for the given image and VQA asks the model to generate answer for the given image-question pair. For image captioning, we choose COCO~\cite{chen2015cococap} and Flickr30K~\cite{plummer2015flickr30ke} as benchmarks and report the CIDEr score. For VQA task, we experiment on VQAv2~\cite{antol2015vqav2} and OK-VQA~\cite{marino2019ok}. 
As shown in Table~\ref{tab:vl}, \textbf{COMM} achieves state-of-the-art performance on image captioning task, \emph{i.e.}, 88.2 CIDEr score on Flickr30K and 132.7 CIDEr score on COCO, even outperforms previous SOTA models with much more parameters (\emph{e.g.}, Shikra-13B with 13B parameters) or much more training data (\emph{e.g.}, Qwen with 1.4B data). For VQA task, our model also shows significant advantages compared to other MLLMs. On VQAv2 val, dev and std, our model achieves 79.05, 81.04 and 81.17 accuracy respectively, which surpasses recent proposed Shikra with the same training data and procedure by a large margin, demonstrating the effectiveness of merging visual embeddings of DINOv2 and CLIP for enhancing visual capabilities. Besides, our \textbf{COMM} model outperforms Qwen with 1.54 and 0.58 accuracy gain on VQAv2 dev and OK-VQA respectively with less VQA training data, \emph{i.e.}, we use 0.6M and Qwen with 3.6M. Training with more VQA data might further improve performance and we leave it as future work.
\subsection{Ablation Study}
\label{sec:mae}
\textbf{Ablation on the MLP of DINOv2.} We conduct ablation study on the design of the MLP module in DINOv2 for aligning visual and text embedding space. 
We ablate on the number and the expanding ratio of MLP module.
Table~\ref{tab:mlp} shows that increasing the number of MLP to 2 can evidently improve performance, demonstrate the effectiveness of using a more powerful network to align the vision only model DINOv2 to the word embedding space. However, increasing the number beyond 2 suffers the degraded performance. For the expanding ratio, increasing to 8 can improve performance, while increasing to 16 does not achieve significant performance gain. Moreover, we experiment with one linear layer, which suffers severe performance degradation. Thus, non-linear MLP is necessary for aligning the features of vision-only DINOv2 to the word embedding space.

\textbf{Ablation on the visual model of MAE and DeiT.} As shown in Table~\ref{tab:deit}, MAE and DeiT suffers from evident performance degradation. For one thing, the visual features of MAE lack sufficient semantic information for global or regional understanding. For another, the supervised training of DeiT is so strong that it learns specialized visual space, making it difficult to align with the word embedding space. 
\begin{table}[!t]
\centering
\setlength{\tabcolsep}{2pt}
\caption{Ablation study on the number and expanding ratio of MLP module. Experiments are conducted on referring expression comprehension and object hallucination benchmark on Random (R), Adversarial (A), and Popular (P).
}
\label{tab:mlp}
\resizebox{\textwidth}{!}{
\begin{tabular}{l|ccccccccc}
\toprule
\multirow{2}{*}{Visual Model}  & \multicolumn{3}{c}{RefCOCO+} & \multicolumn{2}{c}{RefCOCOg}& \multicolumn{3}{c}{RefCOCO}  &POPE\\
& test-A & test-B &val& test-u&  val-u  &  test-A & test-B &val  &A/P/R\\
 \cmidrule(lr){1-10}
DINOv2 w/ MLP Ratio 4
& 75.3 &59.3&67.0&73.0&71.8&84.4&74.1&79.6&80.3/84.2/85.5  \\
\cmidrule(lr){1-10}
DINOv2 w/ 2MLP Ratio 4
& \textbf{77.5}& 	\textbf{60.3} &69.2&	\textbf{74.6} &\textbf{74.7} &\textbf{86.5} &75.3 &81.4 &\textbf{82.4}/84.5/86.2 \\
DINOv2 w/ 4MLP Ratio 4
& 53.7& 	34.4 &45.3&	49.0 &48.8&	65.4 &48.0 &57.9 &79.2/82.9/84.6 \\
DINOv2 w/ 8MLP Ratio 4
& 8.2 & 	6.5 & 	7.4 & 	6.8 & 6.7 & 14.8 & 12.9 & 14.9& 56.0/55.3/59.0\\ 
\cmidrule(lr){1-10}
DINOv2 w/ MLP Ratio 8
&77.4 &59.9 &\textbf{69.7} &73.7 &73.3 &85.7&74.1 &80.9 &81.5/\textbf{85.8}/\textbf{86.7}\\
DINOv2 w/ MLP Ratio 16
&76.2 &60.2 &\textbf{69.7} &74.5&74.6&85.7 &\textbf{75.5} &\textbf{81.5} &80.4/83.7/85.7\\ 
\cmidrule(lr){1-10}
DINOv2 w/ Linear
&61.8& 	48.8 &	55.1 &	64.1 &	62.9 &	76.5 &	67.0 &	71.9& 	75.6/79.3/83.7 \\

\bottomrule
\end{tabular}%
}
\end{table}
\begin{table}[!t]
\centering
\setlength{\tabcolsep}{5.3pt}
\caption{Comparison of the visual model using CLIP, DINOv2 with our multi-level feature merging (MFM), MAE and DeiT. MAE-20 denotes using the features output by the 20-th layer of MAE. DeiT-20 denotes using the features output by 20-th layer.
}
\vspace{-0.1cm}
\label{tab:deit}
\scalebox{1}{
\resizebox{\textwidth}{!}{%
\begin{tabular}{l|ccccccccc}
\toprule
\multirow{2}{*}{Visual Model}  & \multicolumn{3}{c}{RefCOCO+} & \multicolumn{2}{c}{RefCOCOg}& \multicolumn{3}{c}{RefCOCO}  &POPE\\
& test-A & test-B &val& test-u&  val-u  &  test-A & test-B &val  &A/P/R\\
\cmidrule(lr){1-10}
CLIP w/ MFM&  73.7&53.8 &64.3 &69.1 &70.3&83.8 &68.4&76.4&\textbf{80.7}/\textbf{84.2}/\textbf{85.8} 
 \\
 \cmidrule(lr){1-10}
DINOv2 w/ MFM
& \textbf{75.3} &\textbf{59.3}&\textbf{67.0}&\textbf{73.0}&\textbf{71.8}&\textbf{84.4}&\textbf{74.1}&\textbf{79.6}&80.3/\textbf{84.2}/85.5  \\
\cmidrule(lr){1-10}
MAE-20&64.7&	49.4&	56.8	&63.7&	62.8&	77.9&	68.6	&73.6	&66.8/71.1/76.7\\
MAE-22&65.9& 	50.0 &	58.5 &	64.2 &	63.2& 	79.3& 	69.8& 	74.9 &	68.0/71.2/77.5\\ 
\cmidrule(lr){1-10}
DeiT-20&18.4&	13.0&	15.9&	17.0	&16.2	&29.0&	21.6&	25.7&	66.2/69.6/77.9\\
DeiT-22&25.3&	15.4&	19.4&	22.6&	21.8	&36.9&	25.3&	32.0&	67.9/71.6/78.7\\
\bottomrule
\end{tabular}%
}}
\vspace{-0.15cm}
\end{table}
\vspace{-0.12cm}
\subsection{Demonstrations}
\begin{figure}[h!]
\renewcommand{\baselinestretch}{1.0}
\centering
\includegraphics[width=1\columnwidth]{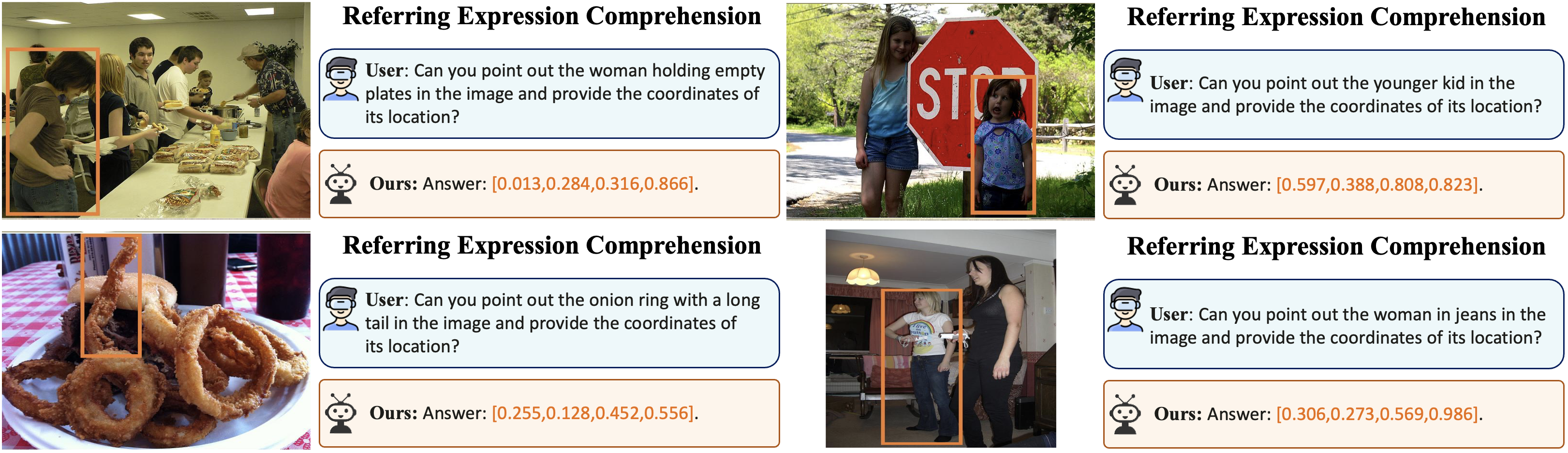}
\vspace{-0.5cm}
\caption{Referring Expression Comprehension (REC) using our \textbf{COMM}-7B. The task intends to localize a target object in an image described by a referring expression.}
\label{fig:caserec}
\vspace{-0.5cm}
\end{figure}

\begin{figure}[!h]
\renewcommand{\baselinestretch}{1.0}
\centering
\includegraphics[width=1\columnwidth]{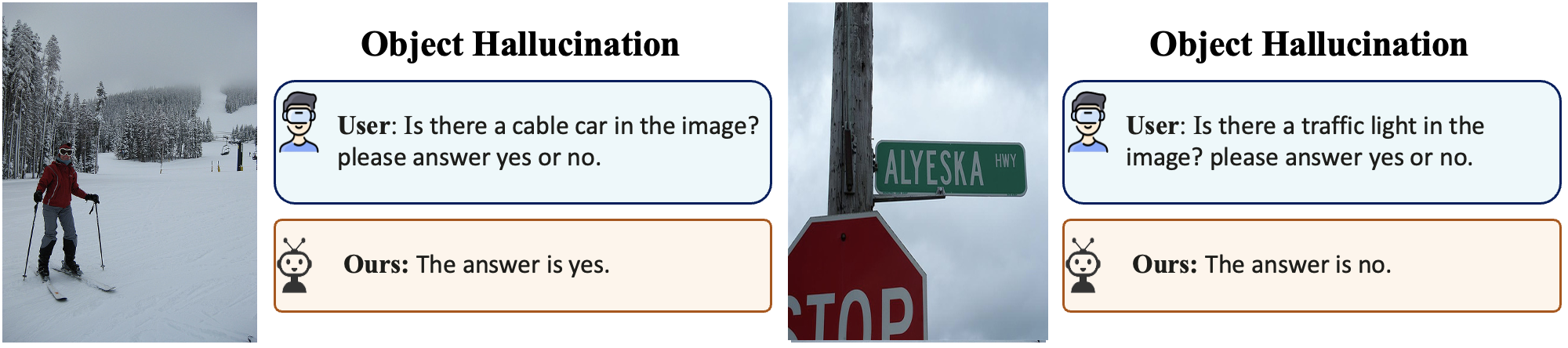}
\vspace{-0.15cm}
\caption{Object hallucination using our \textbf{COMM}-7B. This task aims to evaluate the robustness to object hallucination, \emph{i.e.}, answer yes or no for the existence of questioned object.}
\label{fig:caseobj}
\vspace{-0.5cm}
\end{figure}

\begin{figure}[!t]
\renewcommand{\baselinestretch}{1.0}
\centering
\includegraphics[width=0.9\columnwidth]{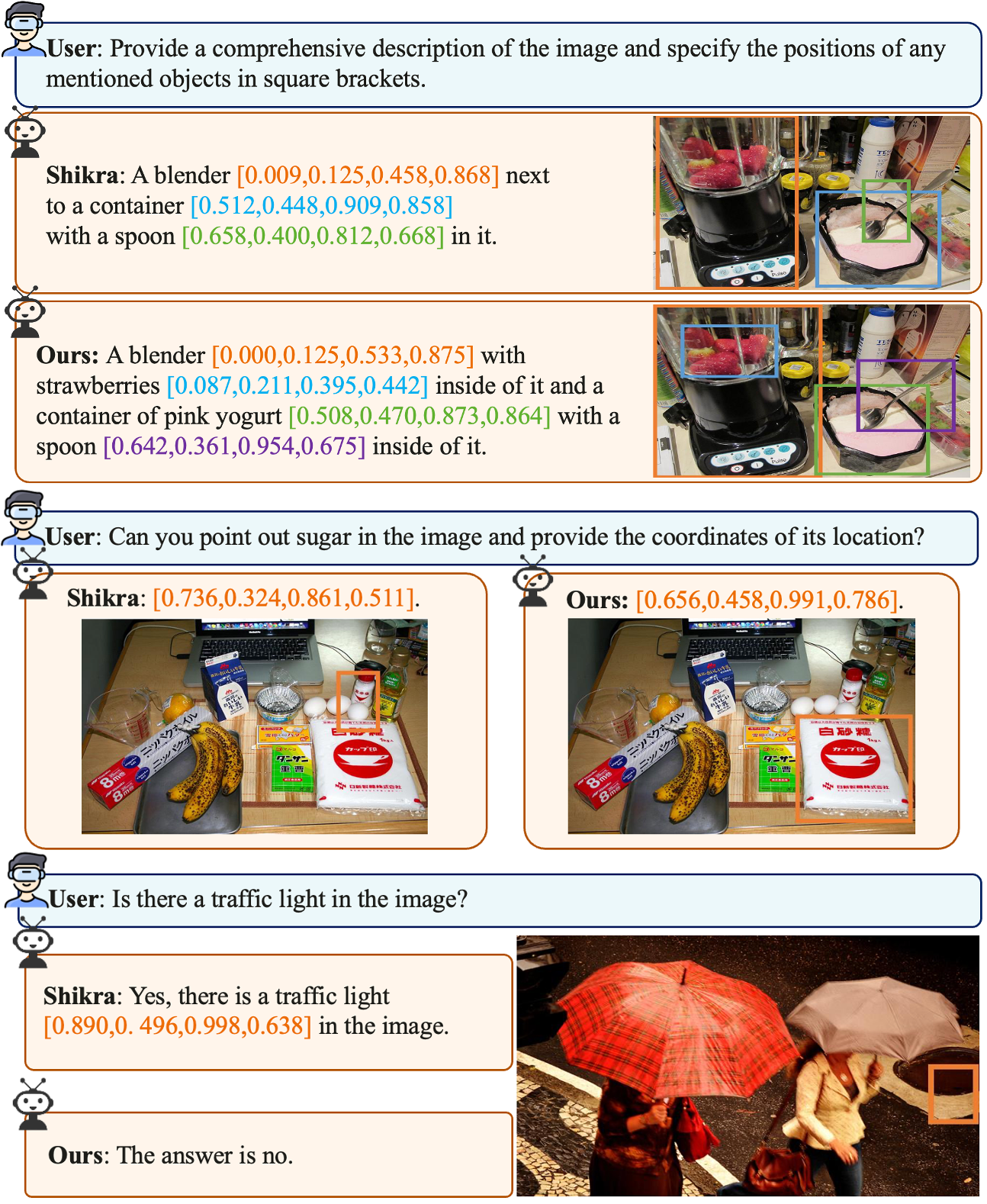}
\vspace{-0.06cm}
\caption{Qualitative comparison between Shikra with its official checkpoint and our \textbf{COMM}.}
\label{fig:case}
\vspace{-0.5cm}
\end{figure}

As shown in Fig.~\ref{fig:case}, our \textbf{COMM} model exhibits a multitude of promising capabilities including visual grounding, fine-grained region understanding and robustness to object hallucination. The first example showcases our strong fine-grained perception ability, which identifies implicit strawberries in a blender. The second example exhibits our strong visual grounding ability to successfully locates the object of sugar. The third case demonstrates our robustness to object hallucination. In contrast, Shikra fails on these challenging cases, showing the superior capabilities of our model. We provide additional demonstrations of our \textbf{COMM} model in this section to demonstrate a multitude of promising capabilities including visual grounding, fine-grained region understanding and robustness to object hallucination. For instance, we showcase Referring Expression Comprehension in Fig.~\ref{fig:caserec} and Object Hallucination in Fig.~\ref{fig:caseobj}.
\vspace{-0.2cm}
\section{Conclusion}
This paper presented an extensive investigation into the efficacy of different visual models when employed as the visual branch in MLLMs. Through a systematic analysis, we highlight the significance of shallow layer features, which capture low-level details that prove beneficial for grounding and positioning tasks. Furthermore, we recognize the potential of the vision-only model DINOv2, which leverages its inherent fine-grained pixel-level information for enhanced fine-grained perception in MLLMs when combined with an MLP layer for alignment purposes. Motivated by our analysis, we introduce a fusion approach to combine the visual features obtained from CLIP and DINOv2, thereby further augmenting the visual capabilities and performance of MLLMs. Through qualitative analysis and extensive quantitative experiments, we demonstrate the effectiveness of our proposed method, surpassing the performance of existing MLLM models across diverse benchmark datasets. Looking ahead, we encourage future research to explore the integration of more powerful vision models to enhance the capabilities of visual branches in MLLMs. We believe that this avenue of investigation holds the key to unlocking the potential of the next generation of MLLMs.
%
%
\bibliographystyle{splncs04}
\bibliography{main}
\end{document}